\documentclass{article}

\usepackage[preprint]{neurips_2026}

\usepackage[utf8]{inputenc} 
\usepackage[T1]{fontenc}    
\usepackage{hyperref}       
\usepackage{url}            
\usepackage{booktabs}       
\usepackage{amsfonts}       
\usepackage{nicefrac}       
\usepackage{microtype}      
\usepackage{xcolor}         

\usepackage{balance}
\usepackage{graphicx}
\usepackage{amsmath}
\usepackage{algorithm}
\usepackage{algorithmic}
\usepackage{wrapfig} 
\usepackage{listings} 
\usepackage{float}    

\usepackage{floatflt}

\usepackage{amssymb}
\usepackage[utf8]{inputenc} 
\usepackage[T1]{fontenc}    
\usepackage{hyperref}       
\usepackage{url}            
\usepackage{booktabs}       
\usepackage{amsfonts}       
\usepackage{nicefrac}       
\usepackage{microtype}      
\usepackage{xcolor}         
\usepackage{wrapfig}

\usepackage{float,wrapfig}
\usepackage{subcaption}
\usepackage{ulem}

\usepackage{multirow}

\usepackage{balance}
\usepackage{xspace}

\definecolor{citypink}{RGB}{227, 108, 194}
\definecolor{cityblue}{RGB}{128, 159, 225}

\usepackage{amsmath}
\usepackage[capitalize]{cleveref}
\crefname{section}{Sec.}{Secs.}
\Crefname{section}{Section}{Sections}
\Crefname{table}{Table}{Tables}
\crefname{table}{Tab.}{Tabs.}
\renewcommand{\paragraph}[1]{\vspace{0.1em}\noindent\textbf{#1}}

\usepackage{multirow}
\usepackage{xcolor}
\usepackage{colortbl}
\usepackage[most]{tcolorbox}
\usepackage{adjustbox}
\usepackage{pifont}
\usepackage{makecell}
\usepackage{array}
\usepackage{tikz}

\definecolor{my_red}{RGB}{204, 0, 0}

\title{ControlLight: Towards Controllable, Consistent, and Generalizable Low-Light Enhancement}

\author{
    {\normalfont\mdseries
    Yufeng Yang$^{1}$ \quad  Jianzhuang Liu$^{1,\dag}$ \quad Jisheng Chu$^{1}$ \quad Yuqi Peng$^{1}$ \quad
    }
    \\[0.4em]
    Xianfang Zeng$^{2}$ \quad Jiancheng Huang$^{1}$ \quad Shifeng Chen$^{1,\ddag}$ \quad
    \\[0.4em]
    \normalsize
    $^{1}$Shenzhen Institutes of Advanced Technology, Chinese Academy of Sciences \quad $^{2}$Zhejiang University 
    \\[0.4em]
    \textcolor{cityblue}{\normalsize 
    \raisebox{-0.2\height}{\includegraphics[height=0.45cm]{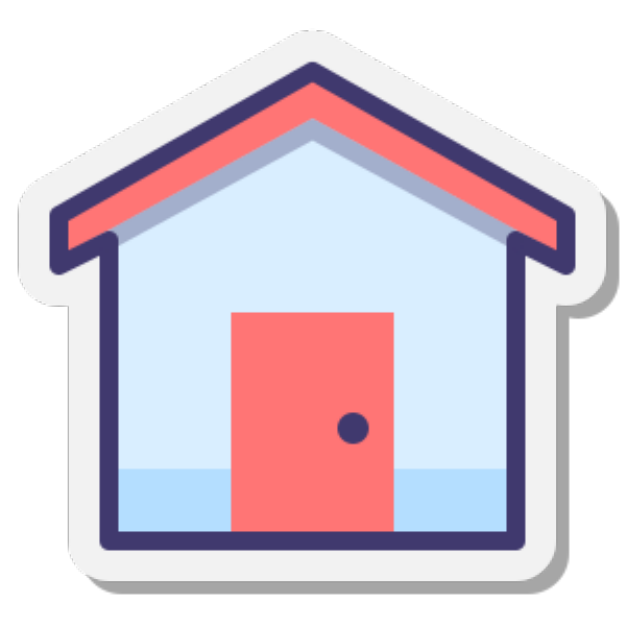}}~\href{https://yfyang007.github.io/ControlLight/}{\textbf{Project Page}}
    \quad
    \raisebox{-0.2\height}{\includegraphics[height=0.45cm]{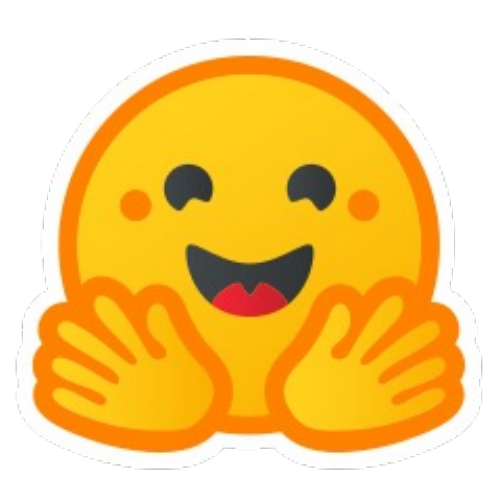}}~\href{https://huggingface.co/ControlLight/ControlLight}{\textbf{Models}}
    \quad
    \raisebox{-0.2\height}{\includegraphics[height=0.45cm]{logos/huggingface_logo.pdf}}~\href{https://huggingface.co/datasets/ControlLight/Light100K}{\textbf{Light100K}}
    \quad
    \raisebox{-0.2\height}{\includegraphics[height=0.45cm]{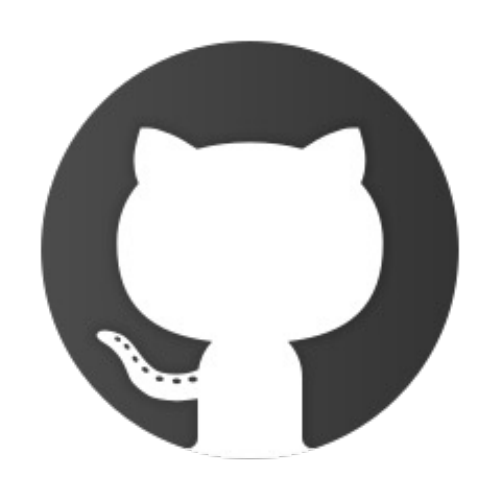}}~\href{https://github.com/yfyang007/ControlLight}{\textbf{Code}}
    }
}

\begin{document}

\maketitle
{\let\thefootnote\relax\footnotetext{\noindent \dag leads this project; \ddag Corresponding authors.}}

\begin{figure*}[h]
    \centering
    \vspace{-20px}
    \includegraphics[width=\linewidth]{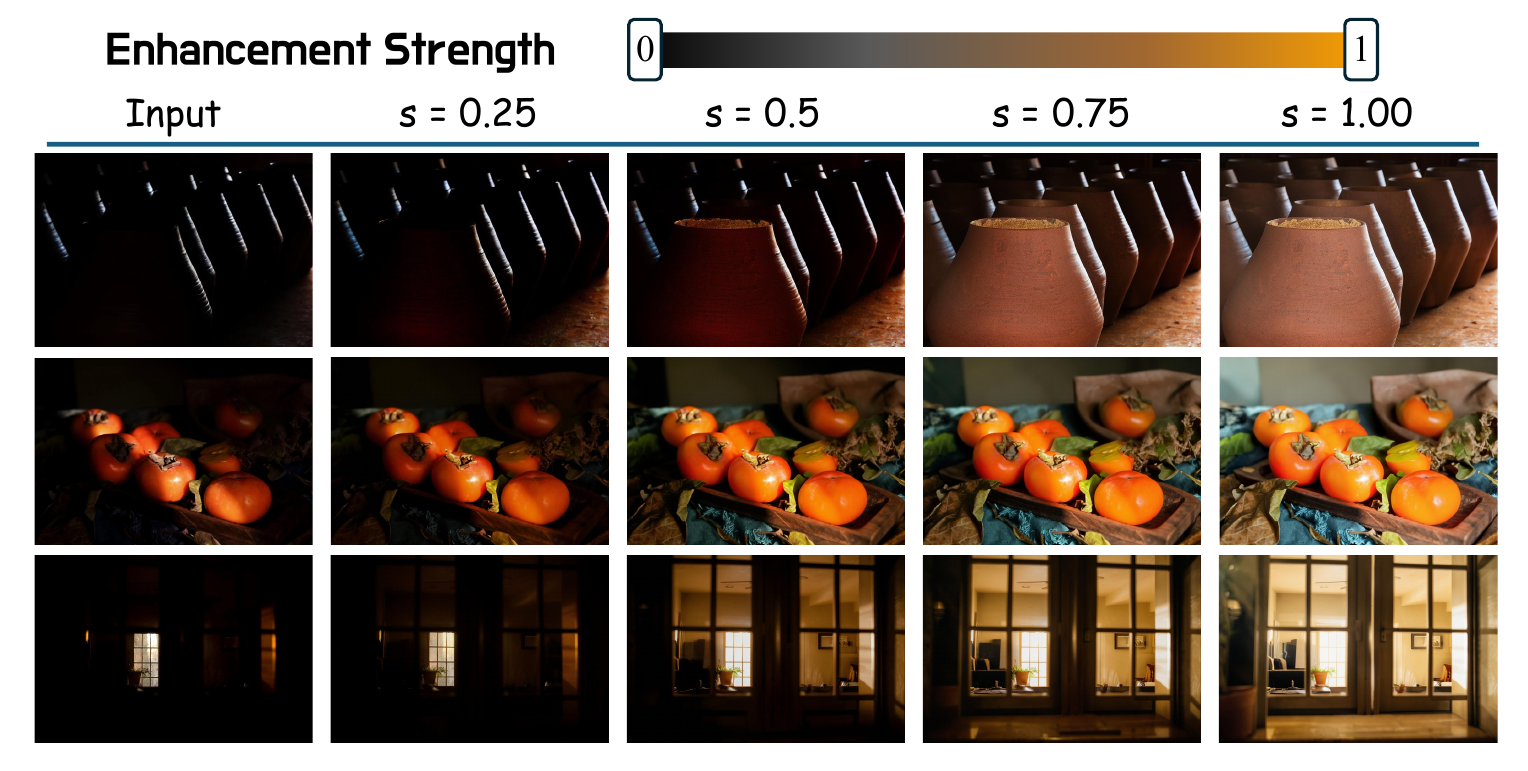}
    \vspace{-0.5em}
    \caption{\small
    Given a low-light input, \textbf{ControlLight} supports continuous adjustment of the enhancement strength from $s=0$ to $s=1$, producing smooth and controllable restoration consistent results across real-world scenes.
    }
    \label{fig:teaser}
\end{figure*}

\begin{abstract}
Existing deep learning-based low-light enhancement methods are typically trained on limited datasets with single enhancement targets, which restricts their generalization ability and controllability in real-world applications. To overcome these limitations, we propose ControlLight, a controllable, consistent, and generalizable framework for low-light enhancement. We first construct a large-scale dataset of real-world degraded images with continuous illumination-strength supervision. To further ensure consistent outputs under different control strengths, we introduce a misalignment-aware weighted flow matching loss that preserves image structure across continuous enhancement strengths. ControlLight allows users to edit real-world degraded low-light images toward satisfactory enhancement results by flexibly controlling the strength while preserving visual consistency and realism. Extensive experiments show that ControlLight achieves state-of-the-art performance against existing low-light enhancement approaches while demonstrating strong continuous controllability and generalization to real-world scenarios.

\end{abstract}


\section{Introduction}\label{sec:intro}

Low-light enhancement aims to recover degraded images captured under low-light conditions by restoring details in dark regions while suppressing noise. With the development of deep learning, many methods~\cite{cai2023retinexformer,chen2018learning,pizer1990contrast,weng2024mamballie,zhang2019kindling,wang2022low,zhou2023pyramid,wang2023ultra} have demonstrated strong capability in low-light image restoration. However, most existing datasets typically provide only a single supervision target for each low-light image, forcing the model to learn a fixed enhancement strength without controllability. This limitation is critical in practical applications, where users often need to freely adjust the enhancement strength according to different images and personal preferences.

Meanwhile, large-scale image editing models~\cite{esser2024scaling,liu2025step1x,wu2025qwenimagetechnicalreport, LongCat-Image}, such as Nano Banana Pro~\cite{team2023gemini} and FLUX.2-klein~\cite{labs2025flux}, have demonstrated strong generalization ability across both high-level and low-level vision tasks. Trained on massive image--text paired data, these models possess powerful generative priors and can recover visually plausible details while largely preserving the overall scene structure, making them promising for low-light enhancement. However, most large image editing models provide a single enhancement strength by giving instructions and may introduce hallucinated textures or structural distortions due to their generative nature, which limits their fine-grained controllability and reliability in practical low-light enhancement scenarios.

To address these issues, we construct \textbf{Light100K}, a continuous low-light enhancement dataset containing real degraded low-light images and structure-consistent pseudo-enhanced targets with different illumination strengths. This dataset provides fine-grained supervision for controllable enhancement.

We further observe that diffusion-generated pseudo targets, despite offering strong appearance supervision, may contain subtle edge misalignment with the input images. Directly applying flow matching to such targets can cause the model to inherit and amplify these offsets, leading to structural artifacts. To mitigate this issue, we propose a Misalignment-Aware Weighted Flow Matching Loss, which down-weights unreliable target-edge regions and encourages structure preservation from the input image.

Based on Light100K and the proposed Misalignment-Aware Weighted Flow Matching Loss, we train \textbf{ControlLight} on FLUX.2-klein-9B with LoRA~\cite{hu2022lora}. By conditioning on the LoRA strength, ControlLight enables continuous and fine-grained low-light enhancement, producing smooth illumination changes while preserving scene structure.

In summary, our contributions are threefold:

\begin{itemize}
    \item We construct \textbf{Light100K}, a continuous low-light enhancement dataset containing training groups, providing fine-grained supervision for controllable low-light enhancement.

    \item We reveal that visually plausible diffusion-generated pseudo pairs can still contain subtle edge misalignment, and propose a \textbf{Misalignment-Aware Weighted Flow Matching Loss} that anchors the enhanced output edges to the input image structure while down-weighting unreliable target-edge regions.

    \item We develop \textbf{ControlLight}, a continuous low-light enhancement model that produces smoothly controllable enhancement results and achieves state-of-the-art performance compared with both continuous and non-continuous low-light enhancement methods.
\end{itemize}

\section{Related Work}\label{sec:related}

\subsection{Low-light Enhancement Methods}
Many deep learning-based low-light enhancement methods~\cite{wang2023ultra,wang2024zero,Feijoo_2025_CVPR} incorporate classical imaging priors, especially Retinex theory~\cite{land1977retinex}, to restore image brightness. With paired datasets such as LOL~\cite{yang2021sparse} and LSRW~\cite{hai2023r2rnet}, these methods learn mappings from low-light inputs to normal-light outputs. EnlightenGAN~\cite{jiang2021enlightengan} learns enhancement from unpaired normal-light images with a GAN-based framework~\cite{zhu2017unpaired}, while Retinexformer~\cite{cai2023retinexformer} uses illumination information to guide a Transformer~\cite{vaswani2017attention}. CIDNet~\cite{yan2025hvi} further revisits brightness restoration from the HSV color space.

These methods are generally trained under fixed supervision and therefore tend to produce results with a single enhancement strength. This limitation makes them less suitable for scenarios where flexible brightness control is required.

To address this issue, several works have investigated controllable low-light enhancement. ReCoRo~\cite{xu2022recoro} adopts GANs to learn enhancement from images with different brightness levels. CLE Diffusion~\cite{yin2023cle} employs a conditional diffusion model that uses brightness alpha blending target images as guidance, enabling controllable enhancement to some extent. Nevertheless, limited by model capacity and the relatively simple interpretation of the training data construction, CLE Diffusion often struggles to generalize to real-world continuous low-light enhancement and may produce noticeable artifacts.

\subsection{Image Editing Methods and Continuous Control}
Large-scale image editing models~\cite{labs2025flux,liu2025step1x,wu2025qwenimagetechnicalreport,LongCat-Image,huang2025diffusion,gao2025seedream,seedream2025seedream,wang2025seededit,seedance2026seedance} have shown strong potential for restoration by leveraging semantic priors learned from massive image--text pairs. However, their generative nature can introduce hallucinations, pixel shifts, and structural deformation, which are undesirable for restoration tasks requiring content consistency. Although high-quality data and consistency reward models~\cite{jiang2026geditbench} can alleviate this issue, existing instruction-based editing methods still lack reliable continuous control.

Recent methods~\cite{baumann2025continuous,gandikota2024concept,parihar2025kontinuous,zarei2025slideredit,sharma2024alchemist,peng2026tara} achieve continuous editing through interpolatable text embeddings, modulation features, or low-rank adaptors, but are limited by scarce continuous supervision. Kontinuous Kontext (KSlider)~\cite{parihar2025kontinuous} synthesizes continuous samples via morphing~\cite{cao2025freemorph}, which is difficult to keep consistent for global restoration tasks. ConceptSlider~\cite{gandikota2024concept} learns controllable LoRA directions, but its control can be unstable without intermediate supervision. To address these limitations, we use Retinex theory to construct continuous pseudo-paired supervision and train a controllable LoRA on FLUX.2-klein-9B. We further propose a Misalignment-Aware Weighted Flow Matching Loss to reduce pixel-level inconsistency during continuous enhancement.

\section{Method}\label{sec:Method}
\subsection{Light100K: Continuous Pseudo-Paired Data Construction}
\label{sec:data}

To address the limited availability of real paired training data, we construct paired data from real-world low-light images rather than relying solely on synthetic degradation generated from traditional single-degradation models.

Specifically, we collect high-quality images from open-source image websites, including Pexels and Pinterest, using low-light-related keywords. To build a high-quality real-world degradation dataset, we conduct low-light semantic and degradation filtering. After filtering, we obtain approximately 30K high-quality low-light images.

\begin{figure*}
    \centering
    \includegraphics[width=\linewidth]{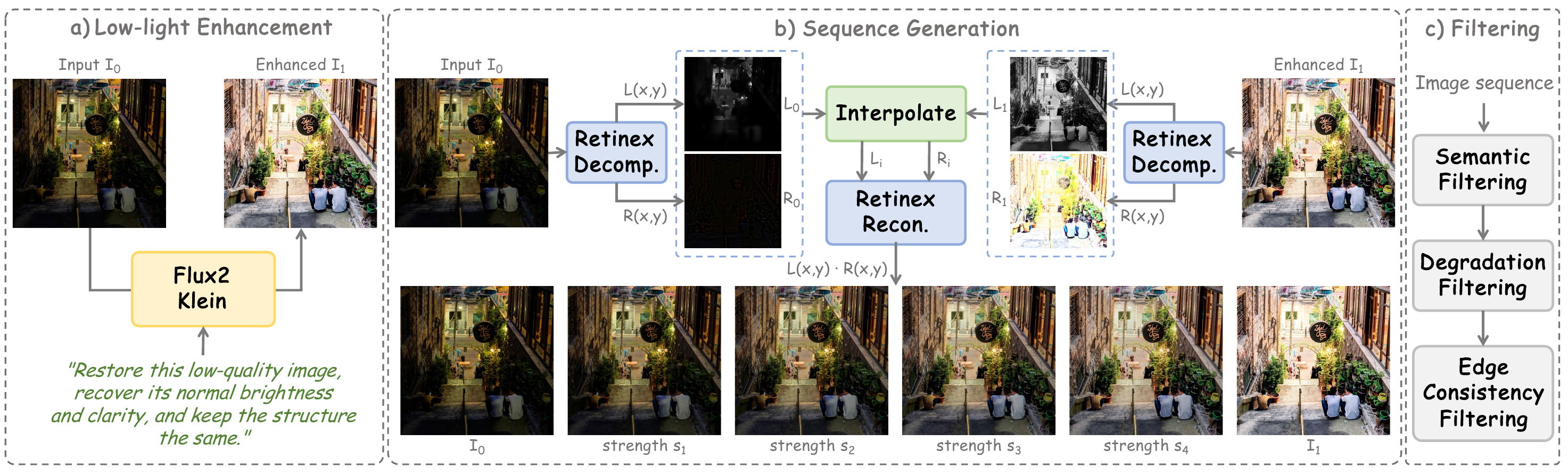}
    \caption{\small
    Main data construction pipeline of Light100K. FLUX.2-klein-9B is used to generate normal-light references from real low-light images. We then apply Retinex-inspired decomposition and selective interpolation to construct highly consistent and continuous pseudo-paired data from each pair $(I_0, I_1)$.
    }
    \label{fig: pipe}
    \vspace{-5mm}
\end{figure*}

Given a low-light image $I_0$, we use a fixed enhancement prompt and the pretrained FLUX.2-klein-9B model to generate its enhanced counterpart $I_1$. To avoid supervision from structurally inconsistent pseudo pairs, we remove severely mismatched samples using an edge-consistency filtering strategy, leaving approximately 20K high-quality paired samples. For each retained pair $(I_0, I_1)$, we further construct a continuous pseudo-paired training group:
\[
    \mathcal{G}
    =
    \{I_0,I_{0.2},I_{0.4},I_{0.6},I_{0.8},I_1\},
\]
where $I_s$ denotes the pseudo ground-truth image at enhancement strength $s \in \{0.2,0.4,0.6,0.8\}$.

A straightforward strategy is alpha blending~\cite{yin2023cle}, i.e., $I_s^{\mathrm{alpha}}=(1-s)I_0+sI_1$. However, direct RGB-space averaging mixes illumination, reflectance, color, and local contrast, making it suboptimal for continuous low-light enhancement where the target should mainly follow a gradual illumination transition.

To construct a more illumination-consistent trajectory, we propose a \textbf{Retinex-inspired interpolation} strategy as shown in Figure~\ref{fig: pipe}. The key idea is to use the Retinex~\cite{land1977retinex} image formation model $I=R\odot L$, where $R$ represents reflectance-related scene content and $L$ represents illumination. Under this model, continuous enhancement is primarily modeled as a transition in the illumination component rather than as a direct interpolation of the whole image appearance.

Specifically, we first convert $I_0$ and $I_1$ from sRGB to linear RGB, and use the same notation for simplicity. We compute their luminance maps by $Y = 0.2126R + 0.7152G + 0.0722B$, and estimate illumination maps using edge-preserving smoothing (bilateral filter): $L_0 = \mathrm{Smooth}(Y_0)$ and $L_1 = \mathrm{Smooth}(Y_1)$, since illumination is assumed to be spatially smooth. The reflectance maps are then estimated according to the Retinex model as $R_0 = I_0 / L_0$ and $R_1 = I_1 / L_1$.

We interpolate only the illumination maps in the log domain rather than $I_0$ and $I_1$ in RGB space:
\begin{equation}
    L_s = \exp\left((1-s)\log(L_0) + s\log(L_1)\right).
    \label{eq:illumination_interp}
\end{equation}
This is equivalent to a multiplicative interpolation $L_s = L_0^{1-s}L_1^s$, which is more consistent with the Retinex assumption than additive image-space averaging. In parallel, we conservatively interpolate the reflectance as $R_s = (1-\beta_s)R_0 + \beta_s R_1$, where $\beta_s = 0.5s$. This design avoids relying only on $R_0$, which may contain amplified low-light noise, while also avoiding excessive dependence on $R_1$, which may inherit artifacts or subtle structural deviations from the diffusion-generated target. The intermediate pseudo-GT is finally reconstructed as:
\begin{equation}
    I_s = \mathrm{clip}(R_s \odot L_s, 0, 1).
    \label{eq:pseudo_gt}
\end{equation}
The reconstructed image is then converted back to sRGB space for training. More details about the data contrsuction pipeline and the Light100K is provided in the Appendix~\ref{app:Data}.

In Figure~\ref{fig:inter}, the direct RGB-space averaging of Alpha blending flattens shadows and textures by weakening local contrast, while the nonlinear illumination transition of Retinex interpolation preserves local shading, scene depth, and contrast variations. Thus, our use of Retinex interpolation provides a more illumination-aware pseudo-GT trajectory for continuously controllable low-light enhancement.

\begin{figure*}[t]
    \centering
    \includegraphics[width=0.9\linewidth]{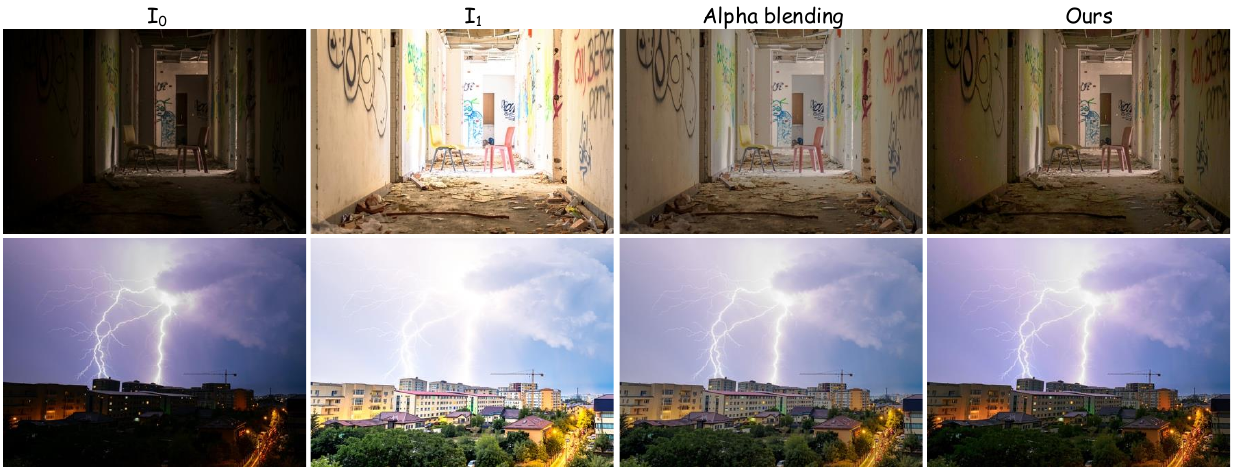}
    \caption{\small
    Visual comparison of intermediate pseudo-GT construction at $s=0.5$. Retinex-based interpolation yields more natural illumination transitions and better local contrast than direct alpha blending, making it better suited for continuous low-light enhancement.
    }
    \vspace{-1.5em}
    \label{fig:inter}
\end{figure*}

\begin{figure*}[t]
    \centering
    \includegraphics[width=0.9\linewidth]{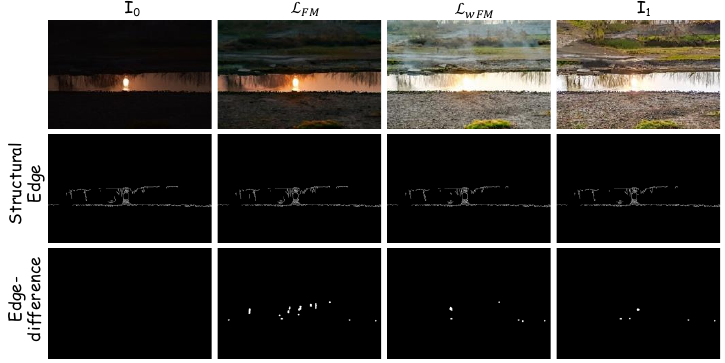}
    \caption{\small
    Visualization of edge misalignment and the effect of weighted flow matching. Compared with standard flow matching, the proposed $L_\mathrm{wFM}$ produces enhanced results with weaker edge-difference responses and better structural alignment to the input.
    }
    \label{fig:wfm}
    \vspace{-1em}
\end{figure*}

\subsection{Misalignment-Aware Weighted Flow Matching}

Although the filtered pseudo pairs $\{I_0,I_1\}$ are visually well aligned, they may still contain subtle pixel-level edge misalignment. Such misalignment is difficult to observe directly in RGB space, as the dominant differences between $I_0$ and $I_1$ mainly arise from brightness and color variations. After normalizing illumination, however, the remaining high-frequency residuals reveal local structural edge discrepancies. As shown in Figure~\ref{fig:wfm}, even a pair that satisfies our matching criterion can still exhibit non-negligible edge differences. When FLUX.2-klein-9B is fine-tuned with the standard flow matching loss~\cite{lipman2022flow,esser2024scaling,peebles2023scalable}, these misaligned edges may be inherited and amplified, leading to visible structural drift in the enhanced output $I_1^{\mathrm{FM}}$. To address this issue, we introduce a misalignment-aware weighted flow matching loss that reduces the supervision strength in unreliable target-edge regions across the continuous pseudo-paired sequence generated from the same degraded image.

To visualize edge misalignment, we employ a structural edge-difference map that focuses on illumination-invariant features. Specifically, we first convert the images to the log-luminance domain and remove slow-varying brightness by subtracting a smoothed version (via a bilateral filter) to isolate the high-pass structural component $H(I)$. We then compute a high-frequency edge response defined as $E(I) = \|\nabla H(I)\|_1$, where $\nabla$ denotes the gradient operator. Finally, the edge-difference map between any two images $A$ and $B$ is calculated as $I_{\text{edge-diff}}(A,B) = |E(A) - E(B)|$. This operation effectively suppresses low-frequency illumination and color discrepancies, ensuring the resulting response primarily reflects local structural misalignments rather than brightness variations.

As shown in Figure~\ref{fig:wfm}, the columns correspond to the input $I_0$, the output trained with $\mathcal{L}_{\mathrm{FM}}$ ($I_1^{\mathrm{FM}}$), our output ($I_1^{\mathrm{wFM}}$), and the pseudo target $I_1$. The second row shows the extracted structural edge maps, while the third row shows the edge-difference maps computed with respect to $I_0$. Compared with standard flow matching, our weighted loss produces fewer edge-difference responses, indicating better preservation of the input structure.

In standard flow matching, given a target image $I_s$ at enhancement strength $s$, we encode it into the latent space as $z_1$, sample a noise latent $z_0$, and construct an intermediate latent $z_t=(1-t)z_0+tz_1$, where $t \in [0, 1]$. The model predicts a velocity field $v_\theta(z_t, I_0, s)$, and the standard objective is:
\begin{equation}
    \mathcal{L}_{\mathrm{FM}} = \left\| v_\theta(z_t, I_0, s) - v^\ast \right\|_2^2,
    \label{eq:standard_fm}
\end{equation}
where $v^\ast = z_1 - z_0$. This objective treats all spatial regions of the pseudo target equally. Therefore, if $I_s$ contains misaligned edges, the model is still encouraged to reproduce those unreliable structures.

We instead assign lower weights to unreliable target-edge regions. For each pseudo target $I_s$, we compute binary edge maps $B_0$ and $B_s$ from $I_0$ and $I_s$, respectively. We then compute the distance transform $D_0$ to the nearest edge pixel in $B_0$. A target edge pixel is regarded as unreliable if it is far from any input edge:
\begin{equation}
    M_s(p) = 1\left[ B_s(p) = 1 \text{ and } D_0(p) > d \right],
    \label{eq:binary_mask}
\end{equation}
where $d$ is a distance threshold. We dilate $M_s$ slightly to cover the neighborhood around the mismatched edge and obtain a soft weight map:
\begin{equation}
    W_s(p) = \mathrm{clip}\left( 1 - \alpha M_s(p), w_{\min}, 1 \right).
    \label{eq:soft_weight}
\end{equation}

Details of the weight map generation and the hyperparameters $d$, $\alpha$, and $w_{\min}$ are provided in Appendix~\ref{app:wfm}.

The image-space weight map $W_s$ is resized to the latent resolution as $\widetilde W_s$, which is then applied over latent spatial locations $u$, to reweight the flow matching objective:
\begin{equation}
    \mathcal{L}_{\mathrm{wFM}} = \frac{\sum_u \widetilde W_s(u) \left\| v_\theta(z_t, I_0, s)(u) - v^\ast(u) \right\|_2^2}{\sum_u \widetilde W_s(u)}.
    \label{eq:weighted_fm}
\end{equation}

Here, $W_s(p)$ remains positive even in unreliable regions, so the model still receives weak appearance supervision but is no longer forced to exactly fit misaligned pseudo-target edges.


\subsection{ControlLight}

\begin{figure*}
    \centering
    \includegraphics[width=0.8\linewidth, height=0.25\textheight]{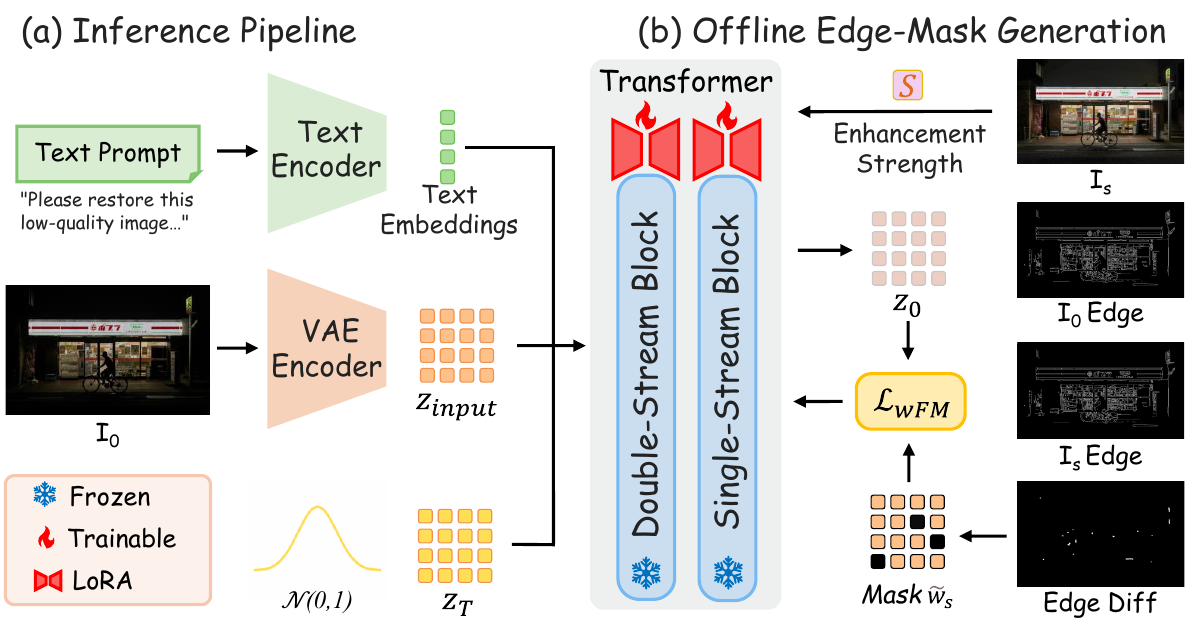}
\caption{\small
Overview of the proposed framework. (a) During training, the low-light input image and a fixed restoration prompt are encoded and fed into FLUX.2-klein, where LoRA is used for efficient fine-tuning. The enhancement strength $s$ modulates both the LoRA scaling factor and the pseudo ground-truth selection. (b) The edge mask is generated offline from input and target edges, producing the weight map $\widetilde{W_s}$. At inference time, $s$ can be set to any value in $[0,1]$.
}
    \label{fig: framework}
    \vspace{-5mm}
\end{figure*}

Given the continuous pseudo-paired dataset (Section~\ref{sec:data}) and the misalignment-aware loss (Eq.~\ref{eq:weighted_fm}), we now describe how $s$ is incorporated into the model. The Retinex formulation $I = R \odot L$ suggests that continuous enhancement is primarily a smooth transition along the illumination axis, approximately linear in some parameter subspace. This motivates using $s$ directly as the LoRA scaling factor:
\begin{equation}
    W' = W + s \cdot AB,
\end{equation}
where $W$ is frozen and $A$, $B$ are learnable low-rank matrices.

This formulation resembles ConceptSlider~\cite{gandikota2024concept}, but the training regimes differ critically. Concept Sliders optimize a LoRA direction via text-guided score matching between opposing prompts, with the scaling factor $s$ applied only at inference time. The linearity of control is assumed but never enforced. In contrast, our $s$ enters the training loop: each $s \in \{0.2, 0.4, 0.6, 0.8, 1.0\}$ is paired with a pseudo ground truth $I_s$, and $\mathcal{L}_{\mathrm{wFM}}$ is computed against that target. The LoRA direction is therefore calibrated against a physically grounded illumination trajectory with per-strength supervision, which is the key reason ControlLight achieves substantially better trajectory smoothness than Concept Sliders (Table~\ref{tab:controllable_core}).

During training, the input image and fixed text prompt are encoded by Flux2-VAE~\cite{labs2025flux} and Qwen3-VL~\cite{qwen3technicalreport}, respectively. Since the prompt remains fixed, the Qwen3-VL text encoder can be offloaded during inference. The weight maps $\widetilde{W}_s$ are precomputed offline. We train at $1024 \times 1024$ resolution with a fixed learning rate of $1\times10^{-4}$ and a global batch size of 16. The LoRA modules contain about 300M trainable parameters. Additional implementation details are provided in Appendix~\ref{app:training}.


\section{Experiments}\label{sec:exp}
\subsection{Quantitative Metrics and Evaluation Protocol}

ControlLight is compared with two baseline groups: low-light enhancement methods and universal continuous image editing methods. For low-light enhancement, we evaluate on five benchmarks: LOL~\cite{yang2021sparse} and LWSR~\cite{hai2023r2rnet} with paired reference, as well as real-world DICM~\cite{lee2013contrast}, LIME~\cite{guo2016lime}, and RealIR-Bench~\cite{yang2026realrestorer} with non-reference. As generative restoration models such as SUPIR~\cite{yu2024scaling} may synthesize perceptually plausible details that are penalized by reference-based metrics like PSNR and SSIM~\cite{wang2004image}, we mainly report non-reference perceptual metrics, including CLIP-IQA~\cite{wang2023exploring}, MUSIQ~\cite{ke2021musiq}, NIQE~\cite{mittal2012making}, and MANIQA~\cite{yang2022maniqa}. To further evaluate Linear Control, we compare ControlLight with universal continuous image editing methods on real-world non-reference test sets, as they can potentially perform continuous low-light enhancement. We assess the smoothness and directionality of the enhancement trajectory using $\delta_{\mathrm{smooth}}$~\cite{parihar2025kontinuous} and CLIP-Dir~\cite{Patashnik_2021_ICCV}, respectively.

\begin{figure*}
    \centering
    \includegraphics[width=\linewidth]{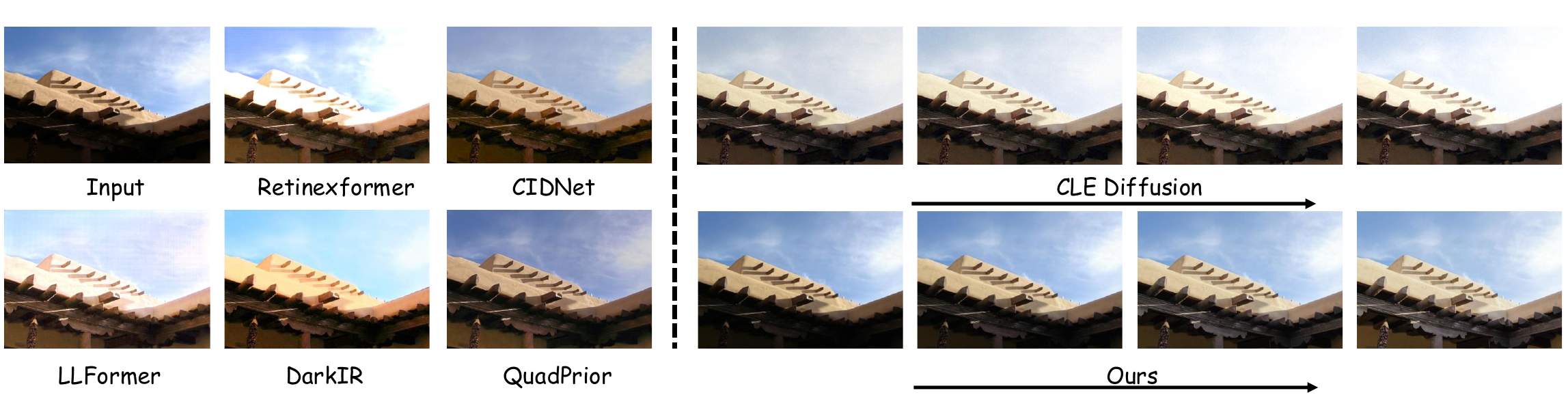}
\caption{\small
Visual comparison on DICM Benchmark. The arrows indicate increasing enhancement strength from left to right. Compared with CLE Diffusion, our method produces smoother and more continuous enhancement transitions while better preserving natural color and scene structure.
}
    \label{fig: comp_dicm}
\end{figure*}

\begin{figure*}
    \centering
    \includegraphics[width=\linewidth]{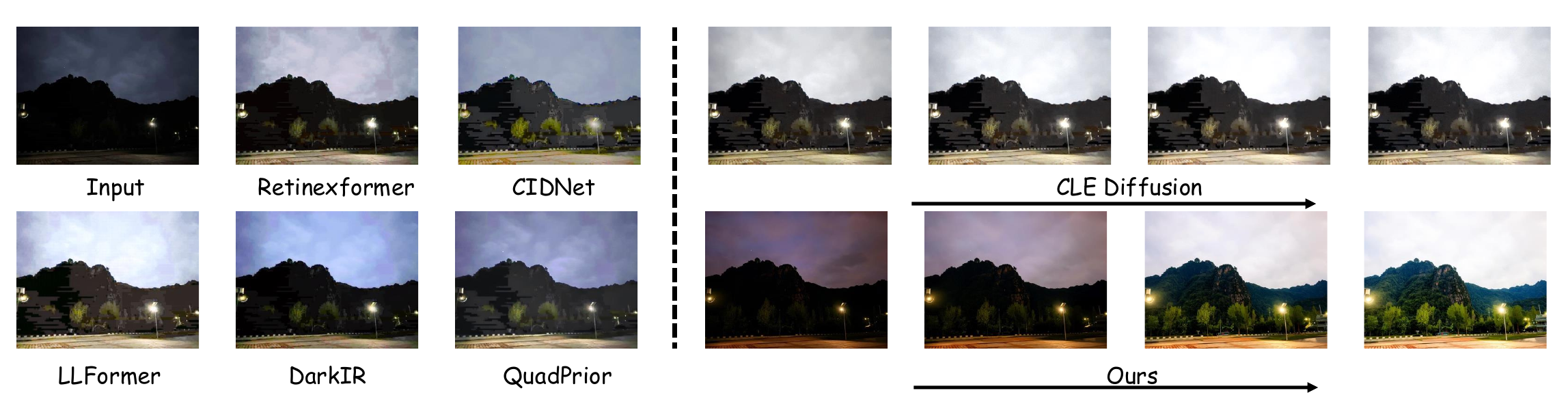}
\caption{\small
Visual comparison on RealIR-Bench. The arrows indicate increasing enhancement strength from left to right. Compared with traditional methods, our method achieves more natural restoration and better preserves scene structure. The arrows indicate increasing enhancement strength, showing that our model provides continuous and controllable low-light enhancement.
}
    \label{fig: comp_realir}
    \vspace{-3mm}
\end{figure*}

\begin{figure*}
    \centering
    \includegraphics[width=0.8\linewidth]{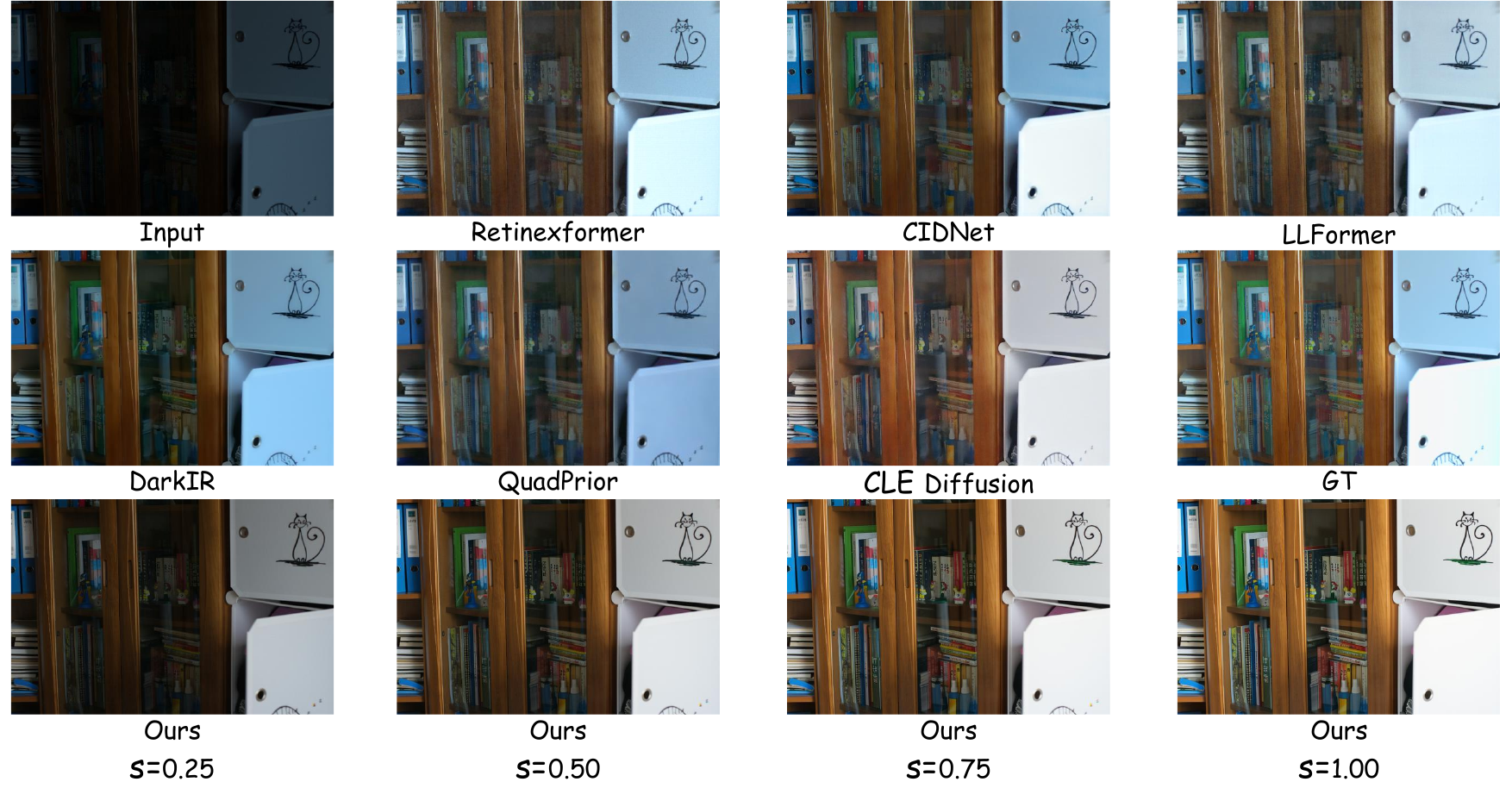}
    \caption{\small
    Visual comparison on LOL-v1 benchmark. Our zero-shot results may deviate from the ground truth in color appearance, but they provide natural visual quality with preserved structures and textures. The outputs at different enhancement strengths $s$ show smooth and approximately linear low-light enhancement control.
    }
    \label{fig:comp_lol}
    \vspace{-0.5em}
\end{figure*}

\subsection{Low-light Enhancement Evaluation}

We compare ControlLight with several state-of-the-art low-light enhancement methods on both paired and unpaired benchmarks. For paired evaluation, we use LOL-v1~\cite{yang2021sparse}, which contains 15 testing images, and the LWSR test set~\cite{hai2023r2rnet}, which contains 50 testing images. For LWSR, we report the average performance over the Huawei and Nikon subsets. The compared methods include Retinexformer~\cite{cai2023retinexformer}, HVI-CIDNet~\cite{yan2025hvi}, LLFormer~\cite{wang2023ultra}, DarkIR~\cite{Feijoo_2025_CVPR}, CLE Diffusion~\cite{yin2023cle}, and QuadPrior~\cite{quadprior}.

Since ControlLight is a continuous enhancement model and does not rely on a single fixed enhancement level, we evaluate it at four enhancement strengths, i.e., $s\in\{0.25,0.50,0.75,1.00\}$, and report the average score. For CLE Diffusion, in paired testing scenarios, the method can use the ground-truth reference to guide result selection. For test sets without ground-truth references, we evaluate CLE Diffusion under the same four-strength setting as ControlLight for a fair comparison.

\begin{table*}[h]
\caption{Quantitative comparison on paired enhancement benchmarks: LOL-v1~\cite{yang2021sparse} and LWSR~\cite{hai2023r2rnet}. The \colorbox{yellow!20}{best} and \colorbox{blue!20}{second-best} results are highlighted with yellow and purple backgrounds, respectively.}
\renewcommand{\arraystretch}{1.2}
\centering
\resizebox{0.95\textwidth}{!}{
\begin{tabular}{l | cccc | cccc}
\toprule
\multirow{2}{*}{Method} 
& \multicolumn{4}{c|}{LOL-v1~\cite{yang2021sparse}}
& \multicolumn{4}{c}{LWSR~\cite{hai2023r2rnet}}
\\
\cmidrule(lr){2-5}\cmidrule(lr){6-9}
& NIQE$\downarrow$ & CLIPIQA$\uparrow$ & MANIQA$\uparrow$ & MUSIQ$\uparrow$ 
& NIQE$\downarrow$ & CLIPIQA$\uparrow$ & MANIQA$\uparrow$ & MUSIQ$\uparrow$ \\
\midrule
Retinexformer~\cite{cai2023retinexformer} 
& \cellcolor{yellow!20}\textbf{3.455} & 0.429 & 0.383 & 63.16 
& \cellcolor{blue!20}3.778 & 0.420 & 0.401 & 58.48 \\
CIDNet~\cite{yan2025hvi} 
& 4.110 & 0.488 & \cellcolor{blue!20}0.511 & \cellcolor{yellow!20}\textbf{71.91} 
& \cellcolor{yellow!20}\textbf{3.708} & 0.415 & 0.387 & 56.25 \\
LLFormer~\cite{wang2023ultra} 
& \cellcolor{blue!20}3.580 & 0.331 & 0.317 & 60.77 
& 3.791 & 0.394 & 0.360 & 57.44 \\
DarkIR~\cite{Feijoo_2025_CVPR} 
& 5.335 & 0.389 & 0.41 & \cellcolor{blue!20}70.69 
& 4.103 & 0.462 & \cellcolor{blue!20}0.431 & \cellcolor{blue!20}64.45 \\

QuadPrior~\cite{quadprior} 
& 5.184 & 0.367 & 0.295 & 58.81 
& 5.045 & 0.345 & 0.358 & 58.91 \\

CLE Diffusion~\cite{yin2023cle} 
& 4.893 & \cellcolor{yellow!20}\textbf{0.581} & 0.435 & 68.84 
& 4.265 & \cellcolor{blue!20}0.491 & 0.388 & 62.63 \\

\midrule
\textbf{ControlLight (Ours)} 
& 4.567 & \cellcolor{blue!20}0.553 & \cellcolor{yellow!20}\textbf{0.512} & 70.20 
& 4.232 & \cellcolor{yellow!20}\textbf{0.589} & \cellcolor{yellow!20}\textbf{0.494} & \cellcolor{yellow!20}\textbf{68.39} \\
\bottomrule
\end{tabular}}
\vspace{-1em}
\label{tab:paired_benchmarks}
\end{table*}

\begin{table*}[h]
\caption{Quantitative comparison on real-world and unpaired datasets: DICM~\cite{lee2013contrast}, LIME~\cite{guo2016lime}, and RealIR-Bench~\cite{yang2026realrestorer}.}
\renewcommand{\arraystretch}{1.2}
\centering
\resizebox{\textwidth}{!}{
\begin{tabular}{l | cccc | cccc | cccc}
\toprule
\multirow{2}{*}{Method} 
& \multicolumn{4}{c|}{DICM~\cite{lee2013contrast}}
& \multicolumn{4}{c|}{LIME~\cite{guo2016lime}}
& \multicolumn{4}{c}{RealIR-Bench~\cite{yang2026realrestorer}}
\\
\cmidrule(lr){2-5}\cmidrule(lr){6-9}\cmidrule(lr){10-13}
& NIQE$\downarrow$ & CLIPIQA$\uparrow$ & MANIQA$\uparrow$ & MUSIQ$\uparrow$ 
& NIQE$\downarrow$ & CLIPIQA$\uparrow$ & MANIQA$\uparrow$ & MUSIQ$\uparrow$ 
& NIQE$\downarrow$ & CLIPIQA$\uparrow$ & MANIQA$\uparrow$ & MUSIQ$\uparrow$ \\
\midrule
Retinexformer~\cite{cai2023retinexformer} 
& 3.962 & 0.377 & 0.291 & 54.27 
& 4.300 & 0.394 & 0.367 & 59.41 
& 4.200 & 0.286 & 0.277 & 52.98 \\
CIDNet~\cite{yan2025hvi} 
& \cellcolor{blue!20}3.657 & \cellcolor{blue!20}0.501 & \cellcolor{blue!20}0.384 & 57.90 
& 4.182 & 0.439 & \cellcolor{blue!20}0.399 & 60.72 
& 4.129 & \cellcolor{blue!20}0.377 & 0.353 & 62.41 \\
LLFormer~\cite{wang2023ultra} 
& 3.943 & 0.435 & 0.274 & 55.03 
& 4.392 & 0.382 & 0.297 & 57.39 
& \cellcolor{blue!20}3.866 & 0.236 & 0.250 & 49.14 \\
DarkIR~\cite{Feijoo_2025_CVPR} 
& 3.869 & 0.463 & 0.345 & 57.44 
& 4.523 & \cellcolor{blue!20}0.441 & 0.373 & \cellcolor{blue!20}61.73 
& 5.097 & 0.374 & \cellcolor{blue!20}0.358 & \cellcolor{blue!20}63.30 \\

QuadPrior~\cite{quadprior} 
& 4.797 & 0.488 & 0.315 & \cellcolor{blue!20}58.21 
& 5.310 & 0.396 & 0.292 & 58.92 
& 4.659 & 0.305 & 0.270 & 51.86 \\

CLE Diffusion~\cite{yin2023cle} 
& 4.368 & 0.390 & 0.218 & 47.38 
& 5.317 & 0.433 & 0.279 & 57.76 
& 3.887 & 0.423 & 0.347 & 60.10 \\

\midrule
\textbf{ControlLight (Ours)} 
& \cellcolor{yellow!20}\textbf{3.522} & \cellcolor{yellow!20}\textbf{0.698} & \cellcolor{yellow!20}\textbf{0.505} & \cellcolor{yellow!20}\textbf{68.22} 
& \cellcolor{yellow!20}\textbf{3.638} & \cellcolor{yellow!20}\textbf{0.576} & \cellcolor{yellow!20}\textbf{0.526} & \cellcolor{yellow!20}\textbf{67.68} 
& \cellcolor{yellow!20}\textbf{3.748} & \cellcolor{yellow!20}\textbf{0.550} & \cellcolor{yellow!20}\textbf{0.491} & \cellcolor{yellow!20}\textbf{67.96} \\
\bottomrule
\end{tabular}}
\label{tab:realworld_benchmarks}
\end{table*}
\begin{table*}[thbp]
\caption{Quantitative comparison of controllable editing performance across three datasets. We focus on the trajectory smoothness ($\delta_{\text{smooth}}$ $\downarrow$) and semantic directional consistency (CLIP-Dir $\uparrow$). To ensure fairness, all the methods are evaluated using aligned four-point control strengths.}
\renewcommand{\arraystretch}{1.2}
\centering
\resizebox{0.8\textwidth}{!}{
\begin{tabular}{l | cc | cc | cc}
\toprule
\multirow{2}{*}{Method} 
& \multicolumn{2}{c|}{RealIR-Bench}
& \multicolumn{2}{c|}{DICM}
& \multicolumn{2}{c}{LIME}
\\
\cmidrule(lr){2-3}\cmidrule(lr){4-5}\cmidrule(lr){6-7}
& $\delta_{\text{smooth}}$ $\downarrow$ & CLIP-Dir $\uparrow$ 
& $\delta_{\text{smooth}}$ $\downarrow$ & CLIP-Dir $\uparrow$ 
& $\delta_{\text{smooth}}$ $\downarrow$ & CLIP-Dir $\uparrow$ \\
\midrule
ConceptSlider~\cite{gandikota2024concept} 
& 0.9237 & -0.0530 
& 0.8700 & 0.3872 
& 0.8589 & 0.0256 \\
AttributeControl~\cite{baumann2025continuous} 
& 0.7262 & \cellcolor{blue!20}0.3520 
& 0.7928 & 0.3593 
& 0.8176 & \cellcolor{blue!20}0.3605 \\
KSlider~\cite{parihar2025kontinuous} 
& \cellcolor{yellow!20}\textbf{0.1956} & 0.0901 
& \cellcolor{blue!20}0.3570 & \cellcolor{blue!20}0.4488 
& \cellcolor{yellow!20}\textbf{0.0485} & 0.0434 \\
SliderEdit~\cite{zarei2025slideredit} 
& 0.3840 & -0.3125 
& 0.4818 & 0.1768 
& 0.3741 & -0.1061 \\
CLE Diffusion~\cite{yin2023cle} 
& 0.7503 & -0.2624 
& 0.7063 & -0.2946 
& 0.6643 & 0.1830 \\
\midrule
\textbf{ControlLight (Ours)} 
& \cellcolor{blue!20}0.2195 & \cellcolor{yellow!20}\textbf{0.9138} 
& \cellcolor{yellow!20}\textbf{0.2382} & \cellcolor{yellow!20}\textbf{0.9012} 
& \cellcolor{blue!20}0.1786 & \cellcolor{yellow!20}\textbf{0.9159} \\
\bottomrule
\end{tabular}}
\vspace{-1em}
\label{tab:controllable_core}
\end{table*}

As shown in Table~\ref{tab:paired_benchmarks} and Table~\ref{tab:realworld_benchmarks}, our method achieves the best results on most metrics among domain-specific methods on paired benchmarks, and consistently outperforms all baselines on real-world benchmarks. This demonstrates its strong generalization capability under real-world degradations. Figure~\ref{fig:comp_lol} further illustrates that our method produces more natural textures and colors. Although such perceptually plausible outputs may deviate from the reference image and slightly affect reference-based metrics (the cat color in Figure~\ref{fig:comp_lol}), they better match real-world visual preference. While due to limited training data and the absence of large-scale generative priors, traditional methods struggle to generalize to realistic low-light degradations, as illustrated in Figure~\ref{fig: comp_realir}. Moreover, our model shows strong linear controllability for low-light enhancement on both paired and real-world benchmarks.

\subsection{Linear Control Evaluation}

Following the evaluation protocol of KSlider~\cite{parihar2025kontinuous}, we report $\delta_{\mathrm{smooth}}$ to measure the smoothness of the continuous enhanment trajectory based on LPIPS feature distances. We also report CLIP-Dir to evaluate whether the enhancement trajectory consistently moves away from dark or underexposed semantics.
We compare with several universal continuous image editing methods, including ConceptSlider~\cite{gandikota2024concept}, AttributeControl~\cite{baumann2025continuous}, KSlider~\cite{parihar2025kontinuous}, SliderEdit~\cite{zarei2025slideredit}, and CLE Diffusion~\cite{yin2023cle}. For a fair comparison, all methods are evaluated at the same four control strengths, $s\in\{0.25,0.50,0.75,1.00\}$, by mapping each method's control variable linearly to this range. As shown in Table~\ref{tab:controllable_core}, our method achieves the highest CLIP-Dir score, demonstrating that its enhancement trajectory is more semantically aligned with the increasing enhancement strength and exhibits stronger linear controllability. More Qualitative Results is provide in the Appendix~\ref{app:extra_experiment}.

\subsection{Ablation Study}

To exmain the effectiveness of our mehtod, we conduct more ablation studies:

\noindent \textbf{Misalignment-Aware Weighted Flow Matching Loss.}
To assess the contribution of the proposed misalignment-aware weighted flow matching loss, we train a baseline model with the standard flow matching objective and evaluate both models on the low-light subset of RealIR-Bench. For consistency evaluation, we adopt LI-LPIPS~\cite{yin2023cle} from CLE Diffusion, an edge-aware and color-normalized perceptual distance that is more stable than the original LPIPS~\cite{zhang2018unreasonable} for measuring continuous-output consistency. We further report non-reference image quality assessment metrics to evaluate perceptual enhancement quality.

\begin{table}[htbp]
\vspace{-3mm}
\caption{Ablation study of $\mathcal{L}_{wFM}$ on the RealIR-Bench. The best results are marked in \textbf{bold}.}
\renewcommand{\arraystretch}{1.2}
\centering
\resizebox{0.75\textwidth}{!}{
\begin{tabular}{l | c | cccc}
\toprule
Ablation &  LI-LPIPS $\downarrow$ & NIQE $\downarrow$ & MANIQA $\uparrow$ & MUSIQ $\uparrow$ & CLIPIQA $\uparrow$ \\
\midrule
$\mathcal{L}_{FM}$ & 0.2237 & 5.6242 & 0.3384 & 55.2252 & 0.5232 \\
$\mathcal{L}_{wFM}$ & \textbf{0.2148} & \textbf{4.5367} & \textbf{0.4180} & \textbf{62.5262} & \textbf{0.6112} \\
\bottomrule
\end{tabular}
}
\label{tab:ablation_wfm}
\end{table}

Table~\ref{tab:ablation_wfm} shows that $\mathcal{L}_{\mathrm{wFM}}$ effectively reduces structural inconsistency while also improving the perceptual quality of the low-light enhancement results.

\noindent \textbf{Data Interpolation Methods.} We conduct a no-reference quality assessment on the five-level interpolation results between Retinex-based interpolation and alpha blending interpolation. As shown in Figure~\ref{fig:inter}, while both methods yield visually plausible results, they differ significantly in their illumination modeling. Specifically, Retinex-based interpolation more faithfully reflects real-world low-light degradation. To evaluate whether Retinex-based interpolation is superior for training, we analyze various no-reference metrics. Table~\ref{tab:ablation_interpolation} in Appendix indicate that Retinex-based interpolation provides richer degradation cues. It exhibits a more pronounced and physically reasonable quality gradient from $ I_1$ to $I_0$, which is essential for the model to effectively learn the enhancement mapping.

\section{Conclusions}\label{sec:conclusion}
We introduced a Retinex-inspired interpolation strategy and a high-quality dataset, Light100K, to facilitate real-world low-light enhancement. To tackle hallucinations and inconsistencies in outputs, we also developed the Misalignment-Aware Weighted Flow Matching Loss with Offline Edge-Mask Generation, which suppresses the effects of edge shifts during training. By fine-tuning the FLUX.2-klein-9B model with LoRA using our proposed $\mathcal{L}_{\mathrm{wFM}}$, ControlLight establishes new state-of-the-art performance. It outperforms existing enhancement and continuous editing methods, delivering superior consistency, controllability, and generalization in real-world scenarios.

\clearpage
{
    \small
    \balance
    \bibliographystyle{plain}
    \normalem
    \bibliography{main}
}


\clearpage
\appendix
\noindent
\textbf{\LARGE Appendix}
\vspace{5ex}
\setcounter{section}{0}
\setcounter{subsection}{0}

\renewcommand{\thesection}{\Alph{section}}
\renewcommand{\thesubsection}{\thesection.\arabic{subsection}}
\makeatletter

\section{Continuous Pseudo-Paired Data Construction Details}
\label{app:Data}
During the construction of Light100K, we first collect high-resolution low-light images from open-source image websites, including Pexels and Pinterest, using low-light-related keywords. We then use the CLIP text encoder~\cite{radford2021learning} to compute the cosine similarity between each image and darkness-related prompts, such as ``a dark photo'', ``underexposed'', and ``low illumination'', in order to filter images with relevant low-light semantics. Next, we employ Qwen3-VL-8B-Instruct~\cite{qwen3technicalreport} to assess the degradation level~\cite{wu2024qbench}, ensuring that the retained images contain sufficient degradation cues for model learning~\cite{jiang2022degrade,rajagopalan2025gendeg}. After semantic and degradation filtering, we obtain 27,529 high-quality low-light images, all with resolutions higher than $1024\times1024$.

We then use FLUX.2-klein-9B to generate restored normal-light references for the collected low-light images. To ensure pairwise structural consistency, we apply Sobel edge detection~\cite{jahne2005digital} to the low-light and restored images and filter out pairs with obvious edge shifts or structural misalignment. This process yields 17,809 high-consistency low-/normal-light image pairs.

Finally, we apply Retinex-inspired interpolation with enhancement strengths $s\in\{0.2,0.4,0.6,0.8\}$ to construct intermediate pseudo targets. The resulting Light100K is a high-quality, real-world, continuous pseudo-paired dataset for controllable low-light enhancement.

\section{Misalignment Analysis and Offline Edge-Mask Generation}
\label{app:wfm}

During the construction of Light100K, subtle visual misalignment may still remain between low-light inputs and their paired enhanced images, even after edge-consistency filtering. Although such misalignment is below the filtering threshold and is often visually negligible, the generative nature of the base model makes it problematic during training. Under the standard flow matching loss,
\[
    \mathcal{L}_{\mathrm{FM}}
    =
    \left\|
    v_\theta(z_t, I_0, s)-v^\ast
    \right\|_2^2 ,
\]
The model is encouraged to fit all target regions equally, which may introduce additional randomness when learning from slightly misaligned pseudo targets and lead to inconsistent outputs.

Our key insight is to preserve the structural edges of the input image while learning the desired illumination enhancement. To this end, we compute illumination-normalized log-luminance representations instead of directly comparing RGB values, so that images with different brightness levels can still share similar structural responses. We then apply a gradient operator to extract the main structural edges.

For each pair in Light100K, we compute a structural edge-difference map and use it to generate a spatial mask that guides flow matching with adaptive weights:
\[
    W_s(p)
    =
    \mathrm{clip}
    \left(
    1-\alpha M_s(p),
    w_{\min},
    1
    \right).
\]
The resulting weight map is resized to the latent resolution as $\widetilde{W}_s$ and used in the weighted flow matching objective:
\[
    \mathcal{L}_{\mathrm{wFM}}
    =
    \frac{
    \sum_u \widetilde W_s(u)
    \left\|
    v_\theta(z_t, I_0, s)(u)-v^\ast(u)
    \right\|_2^2
    }{
    \sum_u \widetilde W_s(u)
    } .
\]

In practice, we set $d=3$ pixels, $\alpha=0.8$, and $w_{\min}=0.2$. To improve training efficiency, all weight maps are generated offline and cached before training.

\section{Implementation Details}
\label{app:training}

During training, we fine-tune only the DiT blocks with LoRA, while freezing the VAE and text encoders. LoRA layers are applied to both the single-stream and double-stream DiT blocks~\cite{peebles2023scalable} with a rank of 64. The bucket resolution is fixed at $1024 \times 1024$, and the global batch size is set to 16. Detailed hyperparameters are provided in Table~\ref{tab:hyperparameters}.  All experiments are conducted on 4 NVIDIA A6000 GPUS.

\begin{table}[htbp]
\centering
\caption{Training hyperparameters for ControlLight fine-tuning.}
\label{tab:hyperparameters}
\renewcommand{\arraystretch}{1.2}
\begin{tabular}{lc}
\toprule
\multicolumn{2}{c}{\textbf{Hyperparameters}} \\
\midrule
LoRA Setting & Rank=64, Alpha=64 \\
Trainable Parameters & 317M \\
Learning Rate & $1\times10^{-4}$ \\
Optimizer & AdamW 8-bit \\
Precision & BFloat16 (BF16) \\
Scheduler & Flow Matching \\
Global Batch Size & 16 \\
Training Steps & 3,000 \\
Resolution & $1024 \times 1024$ \\
\bottomrule
\end{tabular}
\end{table}

\section{More Qualitative Results and Ablation Study Details}
\label{app:extra_experiment}
\subsection{More Qualitative Results}

\begin{figure*}[thbp]
    \centering
    \includegraphics[width=0.98\linewidth]{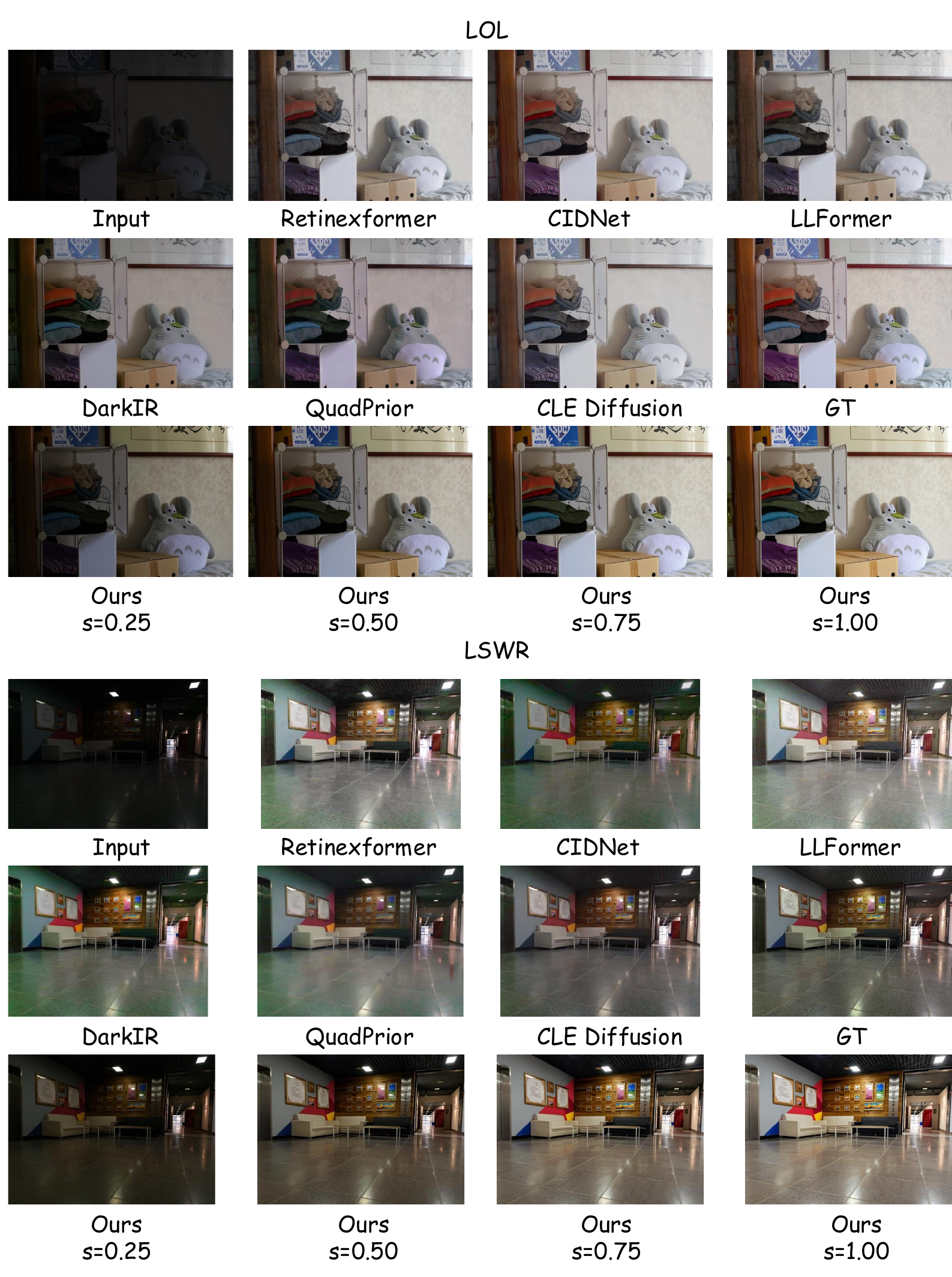}
    \caption{\small Qualitative comparison with state-of-the-art low-light enhancement methods on the paired benchmarks LOL-v1 and LSWR.}
    \label{fig:pairedmore}
\end{figure*}

\begin{figure*}[thbp]
    \centering
    \includegraphics[width=0.98\linewidth]{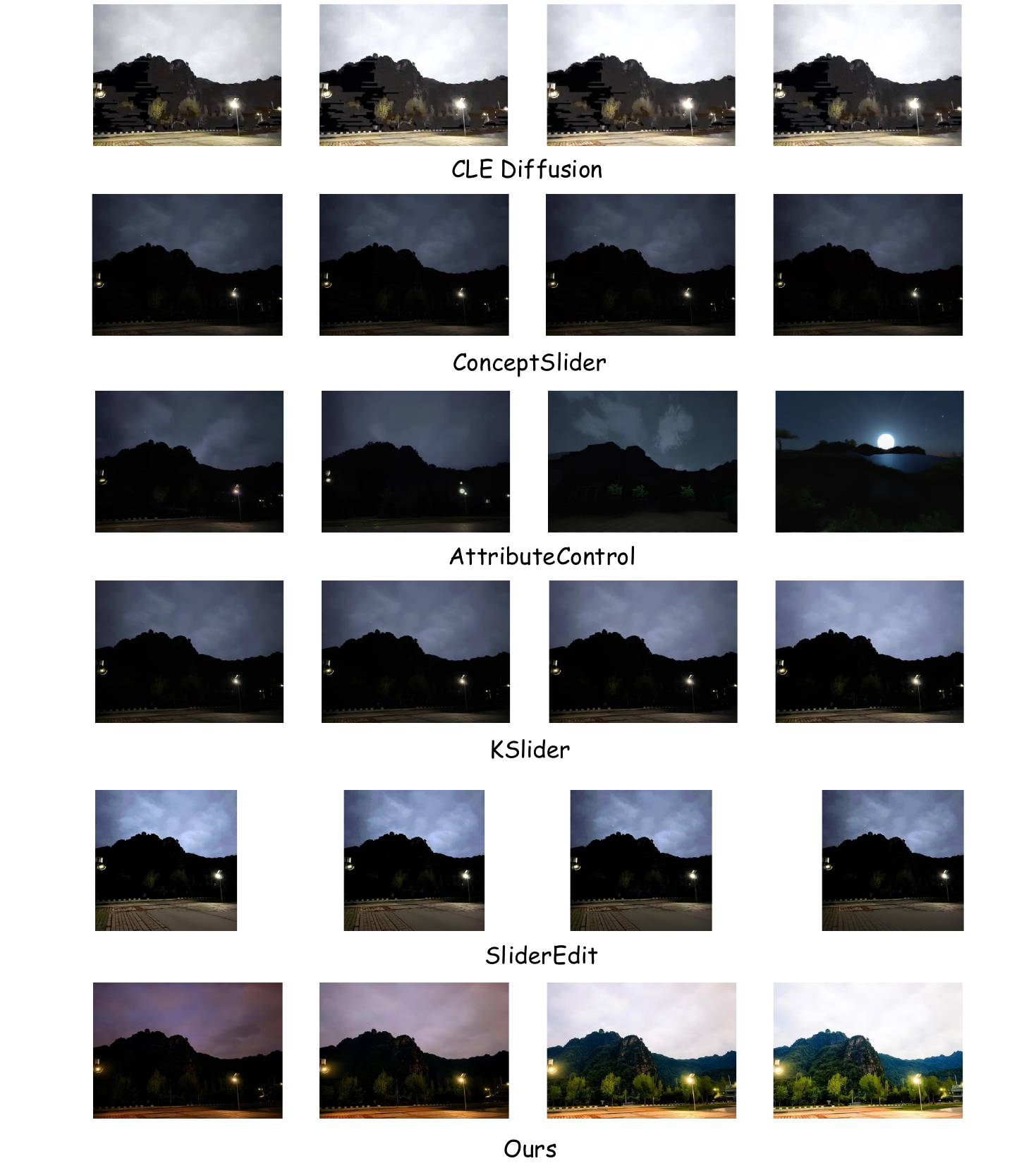}
    \caption{\small Qualitative comparison with universal continuous image editing methods on RealIR-Bench for low-light enhancement. All methods are evaluated under the same four-point control strengths.}
    \label{fig:irmore}
\end{figure*}

Additional qualitative comparisons with low-light enhancement methods and general continuous image editing methods are presented in Figure~\ref{fig:pairedmore} and Figure~\ref{fig:irmore}.

\subsection{More Ablation Study Details}
We report NIQE and MUSIQ to evaluate the image-quality trajectories of different interpolation methods across four intermediate enhancement strengths using 200 randomly sampled images from Light100K. Since the low-light image $I_0$ is expected to have lower perceptual quality than the normal-light image $I_1$, a desirable interpolation method should produce a smooth and monotonic quality transition between them. Table~\ref{tab:ablation_interpolation} shows that Retinex-based interpolation yields more natural image-quality trends and provides richer degradation cues for continuous enhancement learning.
\begin{table}[htbp]
\centering
\footnotesize  
\caption{Ablation study on interpolation strategies for Light100K. Our Retinex-based interpolation preserves the intrinsic degradation at low enhancement levels, whereas Alpha Blending yields artificially high scores that deviate from the low-light distribution.}
\label{tab:ablation_interpolation}
\renewcommand{\arraystretch}{1.2}
\setlength{\tabcolsep}{4pt} 
\begin{tabular}{l | l | c | cccc | c}
\toprule
Metric & Strategy & $I_0$  & $I_{0.2}$ & $I_{0.4}$ & $I_{0.6}$ & $I_{0.8}$ & $I_1$  \\
\midrule
\multirow{2}{*}{NIQE $\downarrow$} & Alpha Blending & 4.588 & 3.931 & 3.561 & 3.356 & 3.461 & 3.695 \\
 & \textbf{Ours} & 4.588 & 4.171 & 3.649 & 3.315 & 3.419 & 3.695 \\
\midrule
\multirow{2}{*}{MUSIQ $\uparrow$} & Alpha Blending & 55.936 & 62.620 & 66.289 & 70.047 & 68.469 & 70.019 \\
 & \textbf{Ours} & 55.936 & 58.780 & 60.889 & 67.626 & 67.716 & 70.019 \\
\bottomrule
\end{tabular}
\vspace{0.5em}
\end{table}

\newpage


\begin{figure*}[thbp]
    \centering
    \includegraphics[width=0.98\linewidth]{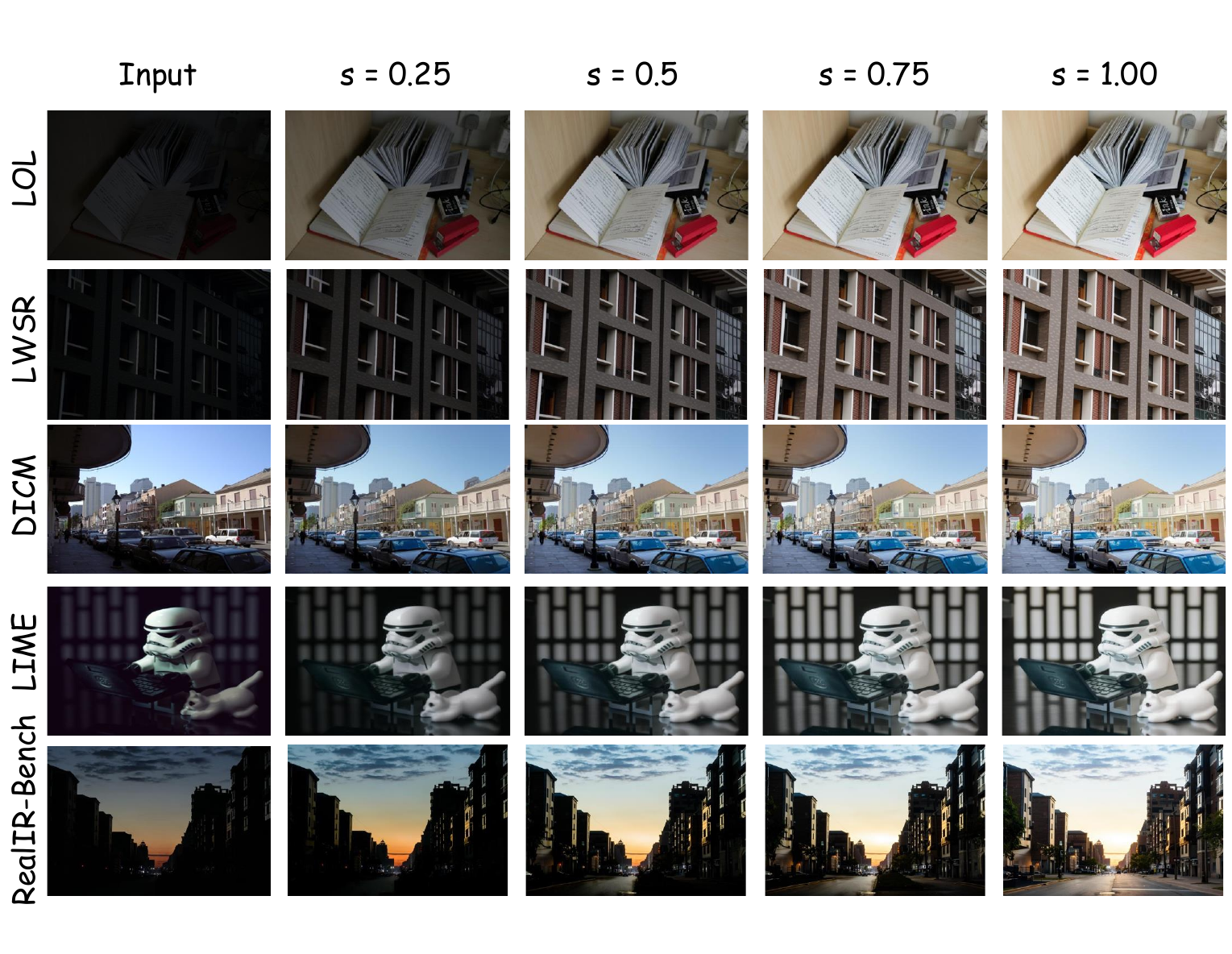}
    \caption{\small Qualitative results of ControlLight on the reported benchmarks under four continuous enhancement strengths.}
    \label{fig:self}
\end{figure*}


\end{document}